\documentclass{article}

\usepackage{arxiv}
\usepackage{amsmath,amsfonts}
\usepackage[utf8]{inputenc} 
\usepackage[T1]{fontenc}    
\usepackage{hyperref}       
\usepackage{url}            
\usepackage{booktabs}       
\usepackage{amsfonts}       
\usepackage{graphicx}
\usepackage{doi}
\usepackage{amssymb}
\usepackage{makecell}
\usepackage{threeparttable}
\usepackage{multirow}
\renewcommand{\headeright}{}

\title{A Novel VAE-DML Fusion Framework for Causal Analysis of Greenwashing in the Mining Industry}

\date{}

\usepackage{authblk}

\newbox{\orcid}\sbox{\orcid}{\includegraphics[scale=0.06]{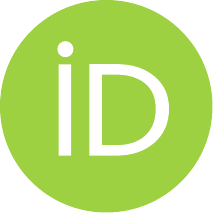}}
\author[1,2]{{\usebox{\orcid}\hspace{1mm}Yuxin Lu}%
}
\author[1,2]{{\usebox{\orcid}\hspace{1mm}Zhen Peng\thanks{\texttt{Corresponding author:pengzhen@cugb.edu.cn}}}%
}
\author[3]{{\usebox{\orcid}\hspace{1mm}Xiqiang Xia\thanks{\texttt{Corresponding author:xqxia@zzu.edu.cn}}}%
}
\author[4]{{\usebox{\orcid}\hspace{1mm}Jie Wang}%
}

\affil[1]{School of Economics and Management, China University of Geosciences (Beijing), Beijing 100083, China}
\affil[2]{MOE Laboratory of Philosophy and Social Sciences for Mineral Resources Security Governance, China University of Geosciences (Beijing), Beijing 100083, China}
\affil[3]{School of Management, Zhengzhou University, Zhengzhou 450001, China}
\affil[4]{Department of Teacher Affairs, Party Committee, China University of Geosciences (Beijing), Beijing 100083, China}

\begin{document}
\maketitle

\begin{abstract}
Amid the global green transition and the advancing ``dual carbon" goals, enterprises within the mining industry chain represent pivotal entities in terms of resource consumption and environmental impact. Their environmental performance bears directly upon regional ecological security and is closely tied to the implementation of national resource strategies as well as the effectiveness of green transformation efforts. Consequently, ensuring the authenticity and reliability of their environmental information disclosure has emerged as a core and pressing issue for promoting sustainable development and achieving strategic national objectives. From a corporate governance perspective, this study focuses on equity balance as a fundamental governance mechanism, systematically investigating its inhibitory effect on greenwashing behavior among mining industry chain enterprises and the underlying pathways through which this effect operates. Methodologically, the paper innovatively introduces a Variational Autoencoder (VAE) and a Double Machine Learning (DML) model, constructing counterfactual scenarios to effectively mitigate endogeneity concerns and thereby precisely identify the causal relationship between equity balance and greenwashing. The findings indicate, first, a significant negative causal relationship between equity balance and corporate greenwashing, confirming its substantive governance effect. Second, this inhibitory effect exhibits notable heterogeneity, manifesting more strongly in western regions, upstream segments of the industrial chain, and industries with high environmental sensitivity. Third, the governance effect demonstrates clear temporal dynamics, with the strongest impact occurring in the current period, followed by a diminishing yet statistically significant lagged effect, and ultimately a stable long-term cumulative influence. Finally, mechanism analysis reveals that equity balance operates through three distinct channels to curb greenwashing: alleviating management performance pressure, enhancing the stability of the executive team, and intensifying media scrutiny.
\end{abstract}

\keywords{Greenwashing\and Mining industry chain\and Double machine learning\and Causal inference\and Variational autoencoder.}

\section{Introduction}
Against the backdrop of increasingly severe global climate change, environmental protection and green sustainable development have become societal imperatives. In 2020, China set the ``dual carbon" goals as a strategic priority to drive the comprehensive green transition of its economy and society. The Sixth Plenary Session of the 19th CPC Central Committee further emphasized that ``protecting the ecological environment means protecting productive forces, and improving the ecological environment means developing productive forces," highlighting the foundational role of ecology in high-quality development. To advance this agenda, China has built a systematic environmental information disclosure framework to reinforce corporate accountability and public oversight. The 19th CPC National Congress initiated the call for mandatory environmental disclosure, leading to the promulgation of regulations such as the Measures for the Management of Legal Disclosure of Enterprise Environmental Informationin 2021, which codified corporate disclosure obligations. The 20th CPC National Congress further emphasized the need to refine the modern environmental governance system, reinforcing the institutional basis for sustained green transformation.

Consequently, enterprises across China face growing pressures regarding environmental information disclosure. This is particularly salient within the mining industry chain—a sector marked by high environmental risk, where activities such as extraction, smelting, and processing carry significant ecological impacts, making it a priority area for both regulatory mandates and public scrutiny \cite{1,2}. At the same time, the industry faces structural constraints, including uneven resource endowments, technological lag, and industrial imbalances \cite{3,4,5}. This tension between stringent environmental requirements and limited capacity for meaningful green investment has heightened incentives for greenwashing—defined as symbolic or ambiguous environmental disclosures used to embellish performance without substantive action \cite{6}. Illustrative cases further highlight this issue: in 2021, Xintiandi Iron Ore Co., Ltd. in Ganzhou illegally expanded production, extracting over 50,000 tons of ore without wastewater treatment facilities, leading to severe ecological harm from open waste piles. Similarly, Daye Nonferrous Metals Group was found operating with nonfunctional exhaust treatment and corroded wastewater systems, resulting in pollutant exceedances. In 2022, Yangguang Sandstone Quarry used superficial measures such as dust nets and artificial foliage in place of substantive ecological restoration. Such practices not only obscure actual environmental performance but may also distort information flows across the industrial chain, misguide policymaking, and ultimately hinder sector-wide green transition \cite{7}. Thus, developing effective mechanisms to curb mining-industry greenwashing represents an urgent academic and practical priority.

Existing literature analyzes corporate greenwashing drivers mainly from two perspectives: internal governance and external constraints. Internally, research focuses on executive characteristics and ownership structure. Studies show that executive environmental awareness and international experience can promote substantive green behavior under ideal conditions \cite{8,9,10}. Yet under high external pressure, such awareness may shift toward symbolic legitimacy, thereby inducing greenwashing \cite{11}. High internal pay gaps also strengthen short-term profit motives, increasing greenwashing tendencies \cite{12}. At the level of ownership structure, research further reveals the complexity of its governance effects. Factors including the short-term orientation of non-green institutional investors \cite{13}, the financial and policy advantages associated with state ownership \cite{14}, and the agency conflicts intensified by controlling shareholders' equity pledges \cite{15} may all contribute to corporate greenwashing. To address these challenges, existing studies indicate that green institutional investors can promote long-term environmental value and mitigate greenwashing through synergistic governance mechanisms \cite{13}, while executive equity incentives help align internal interests and reduce agency costs at the decision-making source \cite{16}. Externally, information monitoring and government regulation serve as key supplements. Media scrutiny raises reputational risks and disciplines corporate behavior \cite{17,18}. Government regulation, on the other hand, relies on punitive deterrence to guide corporate conduct \cite{19,20}. Notably, banking supervisory penalties can curb greenwashing via credit constraints, information signaling, and internal control reinforcement, with effects varying by firm type and penalty nature \cite{21}. However, some policies like environmental taxes may unintentionally encourage greenwashing as firms seek to evade pressure \cite{22}.

Existing research generally assumes that both internal governance and external constraints depend on effective oversight of the supervisors themselves. Whether due to internal executives pursuing short-term gains through greenwashing or external oversight failing due to information asymmetry, both reflect failures in supervision. Specifically, inadequately constrained internal bodies such as boards of directors or supervisors may neglect their duties and tolerate executive opportunism, while insufficiently monitored external supervisors—including regulators and third-party agencies—can exacerbate information asymmetry, rendering oversight nominal. These issues stem from the lack of effective checks on the supervisors themselves. Therefore, establishing feasible mechanisms for balancing power within corporate governance is essential. Equity balance, as a foundational governance structure, constrains controlling shareholders through mutual monitoring among major shareholders \cite{23,24}, with demonstrated effects across conventional governance areas \cite{25,26,27}. This mechanism can strengthen internal oversight bodies' motivation and effectiveness, reducing internal supervision failures, while also limiting corporate information manipulation, thereby easing external supervisors' information asymmetry challenges. However, research on the role of equity balance in mitigating corporate greenwashing remains insufficient. This gap may stem from both theoretical and methodological limitations. Theoretically, equity balance, as a structural variable, tends to influence greenwashing indirectly through mediating mechanisms, making its pathways of effect relatively complex. Furthermore, existing studies have largely focused on its rolein traditional corporate governance issues, while theoretical extensions to emerging deceptive environmental practices such as greenwashing remain underdeveloped. Methodologically, the measurement of equity balance varies widely, and its governance effects are significantly shaped by institutional contexts and industry-specific factors, resulting in limited generalizability of existing findings.

While existing research has examined the governance effects of equity balance and the drivers of greenwashing, few studies have systematically explored their causal mechanisms within the high-environmental-externality and strong-resource-dependency context of the mining industry chain. Given the irreversible and potentially amplified environmental harm that may result from greenwashing in this sector, identifying effective governance tools is particularly crucial. Addressing this gap, this study empirically analyzes the influence of equity balance on corporate greenwashing and its underlying mechanisms, using a sample of 201 listed upstream and downstream firms in the mining industry chain from 2010 to 2022. Its contributions are threefold: First, by situating the analysis in a high-pollution, resource-reliant, and transition-intensive industry setting, it fills avoid in segmented industrial chain governance research and offers industry-level evidence for targeted greenwashing suppression. Second, it constructs a VAE-DML hybrid causal inference framework that combines latent-space counterfactual generation with double machine learning orthogonalization to mitigate sample selection bias and provide robust estimation of causal effects. Third, it integrates behavioral theories such as Prospect Theory and Self-Determination Theory with corporate governance theory, advancing the psychological-motivational understanding of decision-makers and revealing how equity balance restrains greenwashing by alleviating management performance pressure and enhancing internal executive stability while strengthening external media oversight, thereby offering a novel theoretical lens on the causes and governance of corporate greenwashing.

\section{Theoretical Analysis and Research Hypotheses}\label{11}

In mining industry chain enterprises with high environmental sensitivity and capital intensity, concentrated equity enables controlling shareholders to dominate decision-making. Given the environmental impact of activities like mining and smelting, these firms face stringent disclosure requirements and significant long-term compliance costs. This context motivates controlling shareholders to engage in greenwashing—using selective disclosure or misleading publicity to project an eco-friendly image while avoiding substantive investments-particularly when weighing short-term gains against long-term environmental commitments. When control is separated from cash flow rights through mechanisms like pyramid structures or cross-shareholdings \cite{24}, controlling shareholders gain risk-shifting opportunities and enhanced power, often rendering internal oversight bodies such as boards ineffective. This erosion of internal constraints creates space for greenwashing and other opportunistic behaviors driven by personal and risk preferences. From a behavioral perspective, Prospect Theory indicates that decision-makers are loss-averse in uncertain settings—prioritizing loss avoidance over equivalent gains. For mining firms, greenwashing represents a way to evade immediate costs despite uncertain future penalties. Moreover, equity concentration fosters closed social networks that restrict critical information flow, leading controlling shareholders to underestimate the likelihood and severity of exposure. This reinforces short-sighted decision-making, making greenwashing appear as a low-cost strategy to avoid certain expenditures.

In contrast, equity balance serves as an alternative ownership structure that establishes internal checks and balances by introducing multiple major shareholders with comparable stakes. This disperses decision-making power, preventing unilateral control by any single shareholder and thereby curbing greenwashing at its source. The mechanism is especially pertinent in mining industry chain enterprises, where large-scale investments and high asset specificity create significant shareholder exit barriers, aligning shareholder interests with the firm's long-term value. Consequently, shareholders are highly vigilant against greenwashing that may incur regulatory penalties or reputational harm and are motivated to oversee strategic decisions to prevent controlling shareholders' opportunism. Furthermore, environmental risks in the mining sector are systemic, as incidents can trigger chain reactions such as license revocations, fostering a cross-shareholder consensus on environmental compliance that discourages greenwashing. Under this ownership structure, if a shareholder proposes a decision suspected of greenwashing, others can independently evaluate it using their own information channels and employ voting rights to check or jointly oppose it, raising the decision costs and execution risks of such short-sighted actions. Equity balance thus creates an effective internal oversight and game mechanism, increasing the costs and risks tied to greenwashing and other short-term behaviors. This pressure decision-makers to consider broader interests and pursue more long-term, balanced strategies. Hence, the following hypothesis is proposed:

\textbf{H0: Equity balance exerts a significant negative causal effect on the occurrence of greenwashing behavior among upstream and downstream enterprises in the mining industry chain}

Profitability is a key metric used by capital markets to assess corporate performance, with investors typically relying on analyst earnings forecasts to guide decisions. For mining industry chain enterprises, performance volatility is heavily influenced by fluctuations in commodity prices, shifts in environmental policies, and changes in resource reserves. For example, frequent price swings in commodities such as copper and lithium not only affect upstream mining revenues but also impose cost pressures on midstream processors. Differences in environmental policies directly impact compliance costs, while resource reserve uncertainties amplify supply-side risks across the chain, further intensifying performance instability. Given the difficulty in controlling these external factors, performance often falls short of expectations, sparking shareholder dissatisfaction and exposing management to risks such as compensation cuts or dismissal \cite{28}. To mitigate these risks, management may turn to greenwashing as a way to reduce compliance costs and enhance short-term results. This behavior arises not only from economic calculation but also from the impact of external pressures on intrinsic motivation. Self-Determination Theory suggests that individuals possess an innate need for autonomy; when this need is supported, they are more likely to pursue decisions aligned with long-term interests. However, sustained performance pressure from controlling shareholders can undermine managerial autonomy, narrowing focus to short-term performance metrics and increasing the tendency to engage in opportunistic behaviors like greenwashing. This, in turn, diminishes intrinsic motivation to contribute proactively to the organization's long-term value.

The governance structure of equity balance mitigates this dilemma by incorporating diversified shareholder interests. In industries like mining—characterized by technological intensity, long investment horizons, and significant environmental and social risks—shareholder heterogeneity is amplified by sector-specific traits. Financial investors often prioritize short-term cash flow, while strategic or state-owned shareholders focus on long-term resource control and development. Similarly, shareholders diverge on preventive environmental investments: those with high asset specificity and sunk costs tend to support upfront spending to reduce risk, whereas more liquid-asset holders may resist such outlays to preserve short-term returns. This diversity compels management to reconcile competing demands, granting them greater discretionary space and elevating their role from passive executors to autonomous decision-makers. Consequently, performance pressure on management eases, shifting motivation from external coercion toward intrinsic value alignment—thereby weakening incentives for greenwashing aimed at short-term performance embellishment. Moreover, heterogeneous shareholder interests create an internal checks-and-balances system that buffers against short-sighted goals and steers corporate strategy toward equilibrium and long-term value creation. Thus, equity balance restrains greenwashing through dual pathways: enhancing managerial autonomy and intrinsic motivation, while also correcting short-term biases via shareholder counterbalance, thereby aligning decisions with sustainable value.

The stability of the executive team is critical to the long-term development of enterprises, especially in sectors like the mining industry chain that require advanced technical and managerial expertise. A stable team enables the accumulation and transfer of specialized experience and fosters enduring, trust-based relationships with mining communities—key foundations for consistent formulation and implementation of long-term environmental strategies. However, when executives face job insecurity and replacement threats, a psychological threat-rigidity effect often emerges. This narrows cognitive focus, encourages reliance on established solutions, discourages long-term and innovative investments, and shifts priorities toward short-term self-preservation. Job insecurity also intensifies extrinsic motivation, driving an excessive pursuit of external validation and reinforcing short-term decision-making. Prospect Theory further suggests that under conditions of low job security, executives become more loss-averse. Substantive environmental transitions—demanding significant investment and long time horizons—are thus perceived as high-risk, while lower-cost greenwashing, which projects a positive environmental image, aligns more closely with short-term, risk-averse preferences and emerges as a viable strategic choice.

In contrast, a stable executive team fosters deep emotional attachment to the organization through organizational identification, aligning personal success with the firm's long-term development. This encourages executives to prioritize the enterprise's sustainable reputation and growth, making them more inclined to avoid greenwashing that could damage corporate credibility. Existing research confirms that executive stability helps reduce agency costs, shifting managerial focus toward sustainable development rather than short-term gains \cite{29}, and promotes the disclosure of authentic, comprehensive environmental information to build durable competitive advantage \cite{30}. Equity balance further restrains greenwashing by safeguarding executive team stability. On one hand, it prevents arbitrary executive appointments or dismissals by a single shareholder, providing a stable working environment. On the other hand, multi-shareholder negotiation mechanisms reduce turnover risks arising from decision conflicts, while standardized governance protects executives' legitimate rights, lowering voluntary turnover and maintaining team cohesion. A stable tenure environment also strengthens the alignment of executive interests with the firm's long-term goals, reinforcing their commitment to corporate reputation. This, in turn, encourages greater emphasis on long-term environmental compliance and transparency in disclosure, thereby curbing greenwashing at its source.

In mining industry chain enterprises with highly concentrated equity, controlling shareholders can systematically undermine media oversight by leveraging their decision-making dominance. Due to the technical complexity and operational opacity inherent in mineral extraction and smelting, environmental data disclosed by these firms is often difficult for outsiders to verify independently. Controlling shareholders can thus selectively filter or embellish unfavorable environmental information at the source, placing media at a structural disadvantage in oversight. Moreover, when faced with the risk of exposure, controlling shareholders may mobilize corporate resources—such as leveraging advertising partnerships or exerting pressure through affiliated entities—to suppress media scrutiny and obstruct investigative reporting. Under such constraints, media outlets are often forced to soften or abandon critical coverage, significantly weakening their supervisory role. The absence of effective internal checks within these firms allows controlling shareholders to act without internal accountability, further reinforcing their incentive to manipulate environmental information. Together, these factors erode both the information channels and reporting capacity available to the media, severely curtailing their external governance function and creating space for controlling shareholders to engage in greenwashing through controlled disclosure.

In contrast, equity balance enhances external media oversight by introducing diversified shareholder interests. The heterogeneity among major shareholders creates a competitive information environment. If a firm's greenwashing harms a particular shareholder's interests, that shareholder has strong motivation to disclose information to the media to counterbalance other parties, thereby breaking internal information monopolies from within \cite{31}. This transforms the firm–media dynamic from unilateral suppression to multi-party strategic interaction, making it difficult for the firm to fully control information flows. Media gain access to stable, pluralistic sources, improving both the precision and deterrent effect of their oversight. This raises both the probability of greenwashing detection and the certainty of penalties. Research confirms that media exposure can trigger regulatory intervention by revealing environmental violations, imposing reputational costs, and driving firms to abandon greenwashing in favor of genuine environmental improvement \cite{32,33}. In industries like mining, where environmental and social impacts are pronounced, media coverage has a strong social amplification effect, generating significant psychological deterrence. To maintain resource access and social legitimacy, firms tend to internalize environmental pressures as strategic decision criteria, shifting environmental compliance from a cost to a strategic investment. This transition further motivates substantive environmental governance over symbolic greenwashing. Based on this analysis, the following hypothesis is proposed:

\textbf{H1: Equity balance suppresses the occurrence of greenwashing behavior among upstream and downstream enterprises in the mining industry chain by alleviating managerial performance pressure, enhancing the stability of the executive team, and strengthening the effectiveness of media oversight.}

\section{Research Methods and Data Sources}\label{22}
\subsection{Research Methodology}
\subsubsection{Variational Autoencoder (VAE)}
The fundamental challenge in causal inference lies in the unobservability of counterfactuals. For any given unit, only the actual outcome under the received treatment can be observed, while the potential outcome under an alternative treatment remains fundamentally inaccessible. This gives rise to the core missing-data problem in causal discovery. Traditional causal methods primarily rely on observational data and attempt to approximate counterfactuals through statistical adjustment, yet their validity heavily depends on untestable assumptions such as no unmeasured confounding and correct model specification. The Variational Autoencoder is a neural generative model that combines the structure of autoencoders with the principles of variational inference. It learns latent representations of data and generates new samples in an unsupervised manner, with the core objective of learning a probabilistic distribution over latent variables to generate novel samples that resemble the input data \cite{34}. Owing to its distinctive probabilistic generative capability, the Variational Autoencoder can produce high-quality and diverse counterfactual samples, thereby transcending the limitations inherent in observational data. The specific implementation workflow is illustrated in Figure 1.
\begin{figure}[ht]
\centering
  \includegraphics[width=5in, height=2.3in]{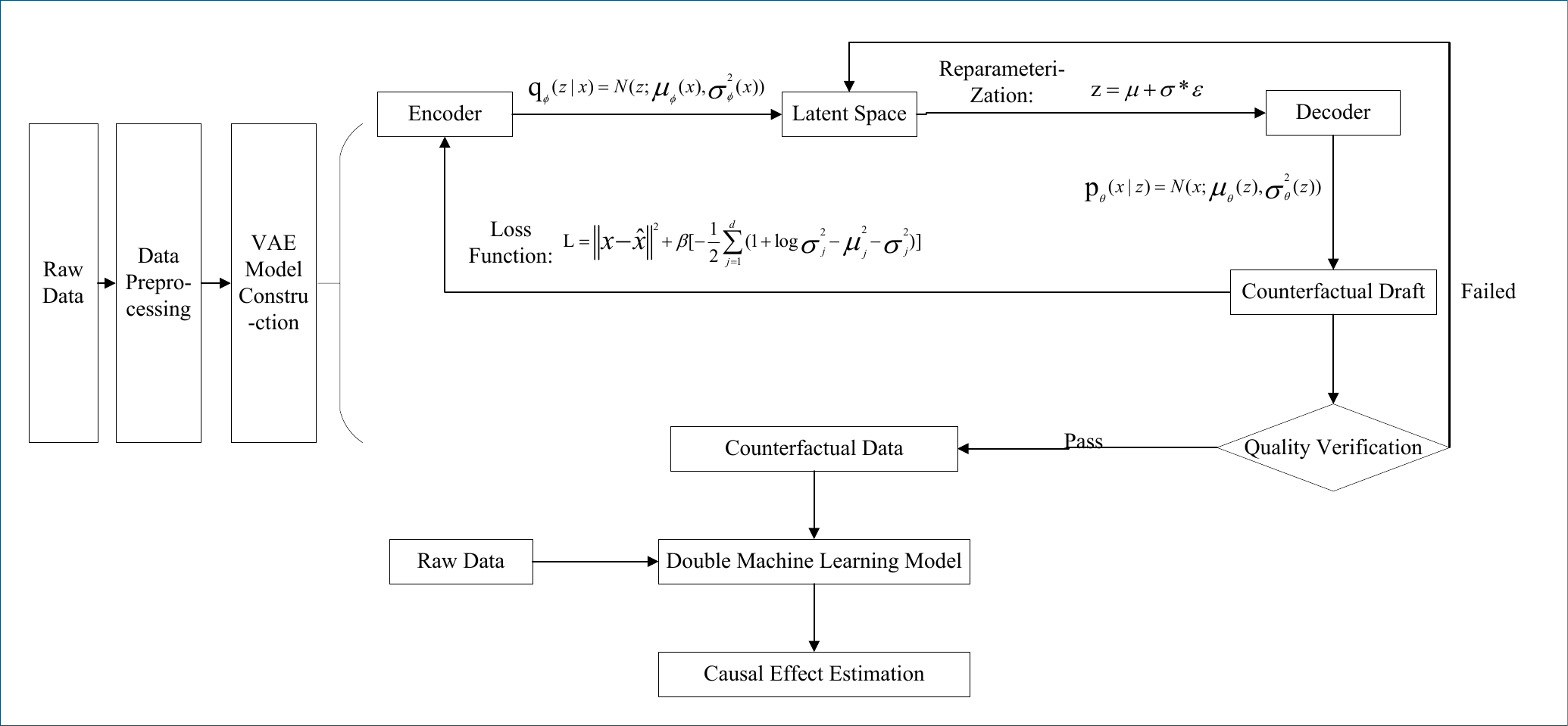}
    \renewcommand\baselinestretch{0.85}
    \vspace{-0.2cm}
\caption{Technical Flow Chart for Counterfactual Data Generation and Causal Effect Estimation Based on VAE.}\label{optimal_price_colume}
\end{figure}

This flowchart illustrates the complete process of counterfactual data generation and causal effect estimation based on Variational Autoencoders (VAE). The process begins with the preprocessing of raw data, followed by the construction of a VAE model comprising an encoder, a latent space, and a decoder. The model is trained using a loss function composed of the Mean Squared Error and the KL divergence weighted by a $\beta$-coefficient. During iterative training, the encoder outputs the mean ($\mu$) and variance ($\sigma^2$) of the latent variables based on the input data. After applying the reparameterization trick to sample latent variables within the latent space, these variables are fed into the decoder to generate an initial draft of counterfactual data. The model is trained through the minimization of the loss function, and network parameters are updated via back propagation. Once the generated counterfactual data satisfies quality verification metrics such as Standardized Mean Difference, Mean Absolute Error, and Mean Squared Error, it is combined with the original data and input into a Double Machine Learning model \cite{35} to perform causal effect estimation.
\subsubsection{Double Machine Learning Model}
Furthermore, how to estimate causal effects from these extended datasets containing both factual and counterfactual data in an unbiased and robust manner becomes a key challenge in causal discovery. While VAE generates counterfactual data through latent space interventions, residual confounding effects that may not be fully removed can persist in such data \cite{36}. Moreover, simply calculating the factual and counterfactual outcomes across all samples yields only an average treatment effect, which cannot precisely capture the heterogeneous causal effects that vary with individual characteristics. In contrast, Double Machine Learning can further remove residual confounding that VAE might have overlooked by analyzing the relationship between the residuals of treatment variables and outcome variables, and it also supports the estimation of conditional average treatment effects \cite{35}. Therefore, this study employs DML for causal discovery, as illustrated in equations (\ref{eq1}) and (\ref{eq2}):
\begin{equation}\label{eq1}
Y=g(D,X)+U, \ \ E(U|X,D)=0,
\end{equation}
\begin{equation}\label{eq2}
D=m(X)+V, \ \ E(V|X)=0,
\end{equation}

In equation (\ref{eq1}), $Y$ is the outcome variable, representing corporate greenwashing behavior; $D$ is the treatment variable, indicating the degree of equity balance; $X$ is a vector of multidimensional control variables, encompassing macroeconomic factors, industry-specific factors, and basic internal corporate characteristics. The functional forms $g(\cdot)$ and $m(\cdot)$ are unknown but can be estimated via machine learning methods; $U$ and $V$ denote the error terms for the outcome and treatment variables, respectively, both with conditional means of zero. The treatment effect is defined as $\theta = E\left[g(D=d,X)-g(D=d_0,X)\right]$, where $d$ and $d_0$ represent different value combinations of the treatment variable.

In equations (\ref{eq1}) and (\ref{eq2}), $g(X)$ and $m(X)$ must be estimated using machine learning algorithms. In this study, the gradient boosting decision tree algorithm is applied in the first stage to estimate $g(D,X)$ and $m(X)$, yielding the corresponding estimates. Subsequently, the residuals $\hat{Y}$ and $\hat{D}$ are calculated. In the second stage, a linear regression model combined with ordinary least squares is employed to estimate the effect of residual $\hat{D}$ on residual $\hat{Y}$, thereby deriving the causal effect estimate $\hat{\theta}$.
\subsection{Variable Selection}
\subsubsection{Outcome Variable}
This study employs the degree of corporate environmental information embellishment as a measure of corporate greenwashing. The degree of environmental information embellishment refers to the discrepancy between the volume of environmental information disclosed by a firm and the substantive amount of environmental information actually disclosed. It captures the extent to which a firm uses textual information to package or mask substantive environmental disclosures.

To comprehensively quantify the textual volume of environmental information disclosure, this study draws on the methodology of Zhang et al. \cite{37}. Specifically, 56 environmental policy documents issued in China in recent years, including the Environmental Protection Law of the People's Republic of China, were collected to extract terms related to green and low-carbon development, ecological environment protection, and pollution control. By integrating these terms with environmental keywords identified in existing literature, a total of 1,749 environmental keywords were compiled after deduplication and manual verification, forming an environmental keyword lexicon. Crawler technology was then employed to retrieve corporate social responsibility reports and environmental reports from the CNINFO website. The content of these reports was converted into text format using the pdfplumber module. The environmental keyword lexicon was incorporated into a custom dictionary to facilitate precise text segmentation via the jieba module. After removing Chinese stop words, the total word frequency of each report and the frequency of environmental keywords extracted based on the lexicon were calculated. The textual disclosure volume was subsequently measured as the ratio of environmental keyword frequency to total word frequency.

The substantive disclosure index system for enterprises is constructed based on the reliability and specificity of environmental information. As shown in Table \ref{tab:substantive_disclosure}, after assigning scores to each indicator, a substantive disclosure score is calculated, with a value range of [0, 36]. Consequently, the degree of environmental information embellishment is measured as the deviation between textual disclosure volume and substantive disclosure volume, as formalized in Equation (\ref{eq3}).
\begin{table}[htbp]
  \centering
  \caption{Indicator System for Corporate Substantive Disclosure}
  \label{tab:substantive_disclosure}\footnotesize
  \vspace{0.1cm}
  \renewcommand{\arraystretch}{1.4}
  \begin{tabular}{|c|c|c|}
    \hline
    \textbf{Category} & \textbf{Substantive Environmental Information Indicators} & \textbf{Scoring Criteria} \\
    \hline
    \multirow{5}{*}{\makecell{Credibility}} & \makecell[c]{Whether the CSR report is prepared in accordance \\with GRI standards} & \makecell[c]{1 if in accordance with GRI,\\ 0 otherwise} \\
    \cline{2-3}
    & ISO14001 certification & 1 if certified, 0 otherwise \\
    \cline{2-3}
    & Whether the auditor is from the Big Four accounting firms & 1 if yes, 0 otherwise \\
    \cline{2-3}
    & Whether the report has been verified by a third-party institution & 1 if verified, 0 otherwise \\
    \cline{2-3}
    & \makecell[c]{Honors or awards obtained by the company in \\environmental protection} & 1 if obtained, 0 otherwise \\
    \hline
    \multirow{10}{*}{\makecell{Substantiveness}} & Compliance of pollutant discharge & 1 if disclosed, 0 otherwise \\
    \cline{2-3}
    & Work safety situation & 1 if disclosed, 0 otherwise \\
    \cline{2-3}
    & \makecell[c]{Disclosure of negative environmental events (3 types): \\ sudden environmental accidents, environmental violations, \\ environmental petition cases} & \makecell[c]{1 if disclosed, 0 otherwise. \\ Value range: [0, 3]} \\
    \cline{2-3}
    & \makecell[c]{Social welfare activities such as environmental \\protection special activities participated by the company} & 1 if disclosed, 0 otherwise \\
    \cline{2-3}
    & \makecell[c]{Establishment of emergency mechanisms for major \\environmental emergencies by the company} & 1 if disclosed, 0 otherwise \\
    \cline{2-3}
    & ``Three Simultaneities" system & 1 if disclosed, 0 otherwise \\
    \cline{2-3}
    & Implementation of cleaner production & 1 if disclosed, 0 otherwise \\
    \cline{2-3}
    & \makecell[c]{Disclosure of pollutant discharge (6 types): wastewater, \\COD, SO$_2$, CO$_2$, smoke and dust, industrial solid waste} & \makecell[c]{0 if no disclosure, \\ 1 if qualitative disclosure, \\ 2 if quantitative disclosure. \\ Value range: [0, 12]} \\
    \cline{2-3}
    & \makecell[c]{Disclosure of pollutant treatment (5 types): waste gas, \\wastewater, smoke and dust, noise and light pollution, solid waste} & \makecell[c]{0 if no disclosure, \\ 1 if qualitative disclosure, \\ 2 if quantitative disclosure. \\ Value range: [0, 10]} \\
    \hline
  \end{tabular}
\end{table}
\begin{equation}\label{eq3}
GW_{i,t}=\frac{mw_{i,t}-\text{mean}(mw_t)}{\text{sd}(mw_{i,t})}-\frac{mr_{i,t}-\text{mean}(mr_t)}{\text{sd}(mr_{i,t})}=mws_{i,t}-mrs_{i,t}.
\end{equation}

In this context, $\text{mean}(mw_t)$ and $\text{mean}(mr_t)$ represent the annual means of textual disclosure volume and substantive disclosure volume, respectively, while $\text{sd}(mw_{i,t})$ and $\text{sd}(mr_{i,t})$ denote the annual standard deviations of textual disclosure volume and substantive disclosure volume. The standardized textual disclosure volume and standardized substantive disclosure volume are expressed as $mws_{i,t}$ and $mrs_{i,t}$, respectively.
\subsubsection{Treatment Variable}
Equity balance refers to a corporate ownership structure in which multiple major shareholders mutually constrain one another, aiming to curb the abuse of control by any single shareholder. Its intensity is typically measured by the ability of other major shareholders to counterbalance the largest shareholder, with values ranging from [0,1). A value closer to 1 indicates stronger balancing capacity, while a value closer to 0 reflects weaker balancing capacity. This study measures the intensity of equity balance using the ratio of the combined shareholding proportion of the second to the tenth largest shareholders to the shareholding proportion of the largest shareholder. To enhance the robustness of the findings, alternative measures—such as ``the ratio of the combined shareholding proportion of the second to the fifth largest shareholders to that of the largest shareholder" and ``the balancing ability of the second largest shareholder relative to the largest shareholder"—will be employed in subsequent robustness tests \cite{24}.
\subsubsection{Control Variables}
This study selects twelve indicators as control variables: Firm Size, Firm Age, Asset-liability Ratio, Return on Total Assets(ROTA), Ownership Concentration, Proportion of Independent Directors, Property Right Nature, Duality of Chairman and General Manager, Executive Incentive, Firm Growth, Industry Competition Degree, Regional Environmental Pollution Degree. Specific details are presented in Table \ref{tab:control_variables}.
\begin{table}[htbp]
  \centering
  \caption{Definition of Control Variables}
  \label{tab:control_variables}
  \footnotesize
  \vspace{0.1cm}
  \renewcommand{\arraystretch}{1.3}
  \begin{tabular}{|c|c|c|}
    \hline
    \textbf{Variable} & \textbf{Definition} & \textbf{Data Source} \\
    \hline
    Firm Size & Natural logarithm of the firm's year-end total assets & \multirow{11}{*}{CSMAR Database} \\
    \cline{1-2}
    Firm Age & Natural logarithm of the firm's establishment years & \\
    \cline{1-2}
    Asset-liability Ratio & Total liabilities at year-end / Total assets at year-end & \\
    \cline{1-2}
    Return on Total Assets (ROTA) & Net profit / Total assets balance & \\
    \cline{1-2}
    Ownership Concentration & Shareholding ratio of the largest shareholder & \\
    \cline{1-2}
    Proportion of Independent Directors & \makecell[c]{Proportion of independent directors on the board \\of directors} & \\
    \cline{1-2}
    Property Right Nature & 1 for state-controlled enterprises, 0 otherwise & \\
    \cline{1-2}
    Duality of Chairman and General Manager & \makecell[c]{1 if the chairman and general manager are the same \\person, 0 otherwise} & \\
    \cline{1-2}
    Executive Incentive & \makecell[c]{Natural logarithm of the total compensation of \\the firm's management} & \\
    \cline{1-2}
    Firm Growth & Operating revenue growth rate & \\
    \cline{1-2}
    Industry Competition Degree & \makecell[c]{Herfindahl index of operating revenue of all firms \\in the industry to which the firm belongs} & \\
    \hline
    Regional Environmental Pollution Degree & \makecell[c]{This paper comprehensively considers industrial \\sulfur dioxide emissions, industrial smoke (dust) \\emissions and industrial wastewater emissions, \\ and uses the entropy weight-TOPSIS method to \\effectively measure the urban environmental \\pollution index.} & \makecell[c]{China Urban\\ Statistical Yearbook, \\ China Environmental\\ Statistical Yearbook, etc.} \\
    \hline
  \end{tabular}
\end{table}
\subsection{Data Sources and Descriptive Statistics}
This study collects 1,743 observations from 201 upstream and downstream enterprises in the mining industry chain during the period 2010-2022. Given the limited volume of raw data and potential bias in the feature distribution, relying solely on raw data may weaken the reliability of identifying the causal relationship between equity balance and greenwashing and constrain the DML model's ability to control for high-dimensional confounders. Therefore, a Variational Autoencoder combined with a resampling approach is employed to generate 1,743 counterfactual data points. Validation using metrics such as Standardized Mean Difference, Mean Squared Error, and Mean Absolute Error confirms that the generated data shows no significant distributional difference from the raw data, effectively avoiding distributional shift. Finally, the raw data and counterfactual data are merged and fed into the DML model to estimate the causal effect of equity balance on corporate greenwashing. The descriptive statistics of the merged dataset are presented in the table below:
\begin{table}[htbp]
  \centering
  \caption{Descriptive Statistics}
  \label{tab:descriptive_statistics}\footnotesize
  \vspace{0.1cm}
  \renewcommand{\arraystretch}{1.2}
  \begin{tabular}{ccccccc}
    \toprule
    \textbf{Variable Name} & \textbf{Sample Size} & \textbf{Mean} & \textbf{Median} & \textbf{Std. Dev.} & \textbf{Min.} & \textbf{Max.} \\
    \midrule
    Greenwashing Degree & 3486 & -0.0060 & -0.0160 & 1.1140 & -3.1538 & 4.6597 \\
    Equity Balance Degree & 3486 & 0.2975 & 0.2098 & 0.2714 & 0.0013 & 1 \\
    Regional Pollution & 3486 & 0.0750 & 0.0618 & 0.0578 & 0.0004 & 0.5163 \\
    Comprehensive HHI & 3486 & 0.1922 & 0.1457 & 0.1110 & 0.0888 & 0.9675 \\
    Firm Size & 3486 & 23.5987 & 23.5567 & 1.2046 & 19.1979 & 28.6365 \\
    Asset-Liability Ratio & 3486 & 0.5223 & 0.5279 & 0.1478 & 0.0156 & 1.3986 \\
    Return on Total Assets (ROTA) & 3486 & 0.0375 & 0.0339 & 0.0600 & -0.6438 & 0.9533 \\
    Firm Growth & 3486 & 0.1635 & 0.1452 & 0.2933 & -0.9913 & 6.1752 \\
    Proportion of Independent Directors & 3486 & 0.3720 & 0.3651 & 0.0397 & 0.2500 & 0.6667 \\
    Duality of Chairman and CEO & 3486 & 0.1216 & 0 & 0.3269 & 0 & 1 \\
    Shareholding Ratio of the Largest Shareholder & 3486 & 0.4212 & 0.4148 & 0.1632 & 0.0339 & 0.8999 \\
    Firm Age & 3486 & 2.9227 & 2.9444 & 0.2574 & 1.3863 & 3.6636 \\
    Executive Incentive & 3486 & 15.3584 & 15.3624 & 0.6964 & 11.3206 & 18.5134 \\
    Property Right Nature & 3486 & 0.6661 & 1 & 0.4717 & 0 & 1 \\
    \bottomrule
  \end{tabular}
\end{table}

\section{Empirical Results Analysis}\label{33}
\subsection{Causal Inference Results}
To ensure the accuracy and robustness of causal inference, this study employs a cross-fitting Double Machine Learning model, with a sample-splitting ratio set at 1:4. Estimations are conducted separately on the original dataset and the merged dataset that incorporates the VAE-generated counterfactual data, in order to systematically identify the causal effect of equity balance on greenwashing behavior.
\subsubsection{Causal Effect Estimation Based on the Original Dataset}
First, this study conducts causal inference on the original dataset, with the results presented in Table \ref{tab:causal_inference_raw}. The analysis reveals that the estimated coefficients of equity balance are consistently negative, indicating a preliminary sign of an inverse relationship between equity balance and corporate greenwashing behavior.
\begin{table}[htbp]
  \centering
  \caption{Causal Inference Results of Raw Data}
  \label{tab:causal_inference_raw}\footnotesize
  \vspace{0.1cm}
  \renewcommand{\arraystretch}{1.2}
  \begin{tabular}{ccccc}
    \toprule
    Variable & (1) & (2) & (3) & (4) \\
    \midrule
    Equity Balance Degree & -0.2946 & -0.2258 & -0.2106 & -0.2318 \\
    & (0.1913) & (0.1884) & (0.1973) & (0.1976) \\
    Linear Terms of Control Variables & YES & YES & YES & YES \\
    Quadratic Terms of Control Variables & NO & YES & YES & YES \\
    Industry Fixed Effects & NO & NO & YES & YES \\
    Year Fixed Effects & NO & NO & NO & YES \\
    Number of Observations & 1743 & 1743 & 1743 & 1743 \\
    \bottomrule
    \multicolumn{5}{l}{\scriptsize Notes: Robust standard errors in parentheses.} \\
  \end{tabular}
\end{table}

However, the robust standard errors in the results shown in Table 4 are relatively large, leading to low estimation precision, and none of the coefficients pass the test of statistical significance. This suggests that inference based solely on the original sample—due to limitations such as a restricted sample size and potential selection bias—may struggle to clearly and robustly identify the causal effect of equity balance on greenwashing behavior.

\subsubsection{Causal Effect Validation Based on the Merged Dataset}
To address the limitations of the original dataset, this study combines the original dataset with the VAE-generated counterfactual dataset and performs causal inference again. The results are presented in Table \ref{tab:causal_inference_pooled}, which shows significant improvements compared to Table \ref{tab:causal_inference_raw}.
\begin{table}[htbp]
  \centering
  \caption{Causal Inference Results of Pooled Data}
  \label{tab:causal_inference_pooled}\footnotesize
  \vspace{0.1cm}
  \renewcommand{\arraystretch}{1.1}
  \begin{tabular}{ccccc}
    \toprule
    Variable & (1) & (2) & (3) & (4) \\
    \midrule
    Equity Balance Degree & -0.3952$^{***}$(0.0844) & -0.4329$^{***}$(0.0976) & -0.3986$^{***}$(0.1089) & -0.3726$^{***}$(0.1036) \\
    Linear Terms of Control Variables & YES & YES & YES & YES \\
    Quadratic Terms of Control Variables & NO & YES & YES & YES \\
    Industry Fixed Effects & NO & NO & YES & YES \\
    Year Fixed Effects & NO & NO & NO & YES \\
    Number of Observations & 3486 & 3486 & 3486 & 3486 \\
    \bottomrule
    \multicolumn{5}{l}{\scriptsize Notes: *** $p<0.01$, robust standard errors in parentheses.} \\
  \end{tabular}
\end{table}

By comparing Table \ref{tab:causal_inference_raw} and Table \ref{tab:causal_inference_pooled}, the following observations can be made. First, the sample size after data merging doubles, contributing to improved estimation efficiency and reduced sampling error. Second, the coefficient for equity balance in Table 5 is not only highly statistically significant at the 1\% level but also exhibits markedly smaller robust standard errors compared to those in Table4, indicating a substantial enhancement in the precision of causal effect estimation. Furthermore, since causal inference must account for different scenarios and possibilities, counterfactual data can construct virtual control scenarios that closely resemble the real samples, effectively filling the data gap corresponding to ``no intervention" in observational data. This significantly strengthens the stability of statistical inference and the credibility of causal effect identification. These results clearly identify a robust negative causal effect of equity balance on greenwashing behavior, supporting hypothesis H0. In summary, the phased causal inference demonstrates that incorporating counterfactual data to construct a merged dataset significantly improves the precision and reliability of causal identification, thereby validating the core conclusion that equity balance has an inhibitory effect on greenwashing behavior in mining enterprises.
\subsection{Robustness Test}
\subsubsection{Winsorization and Model Algorithm Substitution}
To verify the robustness of the conclusions, this study conducts tests from two aspects: data processing and model specification. On one hand, in order to mitigate the influence of extreme values caused by differences in geographic location and economic foundations among enterprises, all variables were subjected to a 1\% bilateral winsorization before reanalysis. On the other hand, to ensure that the findings are not dependent on the model specification and sample splitting ratio, alternative DML model algorithms such as LASSO, XGBOOST, and Random Forest were employed, and sample splitting ratios of 1:2 and 1:7 were used to replace the original 1:4 ratio. The results are presented in Table \ref{tab:robustness_results}.
\begin{table}[htbp]
  \centering
  \caption{Robustness Results}
  \label{tab:robustness_results}\footnotesize
    \vspace{0.1cm}
  \renewcommand{\arraystretch}{1.1}
  \begin{tabular}{ccccccc}
    \toprule
    \multirow{2}{*}{Variable} & \multicolumn{3}{c}{Changing Machine Learning Algorithms} & \multicolumn{2}{c}{Changing Sample Split Ratio} & \multirow{2}{*}{Winsorization} \\
    \cmidrule(lr){2-4} \cmidrule(lr){5-6}
    & LASSO & XGBOOST & Random Forest & 1:2 & 1:7 & \\
    \midrule
    Equity Balance Degree & -0.2733$^*$ & -0.4015$^{***}$ & -0.2165$^{***}$ & -0.3765$^{***}$ & -0.4210$^{***}$ & -0.3412$^{***}$ \\
    & (0.0548) & (0.0973) & (0.0608) & (0.0021) & (0.0031) & (0.0007) \\
    Linear Terms of Control Variables & YES & YES & YES & YES & YES & YES \\
    Quadratic Terms of Control Variables & YES & YES & YES & YES & YES & YES \\
    Year Fixed Effects & YES & YES & YES & YES & YES & YES \\
    Industry Fixed Effects & YES & YES & YES & YES & YES & YES \\
    Number of Observations & 3486 & 3486 & 3486 & 3486 & 3486 & 3486 \\
    \bottomrule
    \multicolumn{7}{l}{\scriptsize Notes: $^*$ $p<0.1$, $^{***}$ $p<0.01$; robust standard errors in parentheses.} \\
  \end{tabular}
\end{table}

As shown in Table \ref{tab:robustness_results}, after applying 1\% bilateral winsorization, the causal effect coefficient of equity balance on greenwashing behavior is -0.3412 and remains highly significant at the 1\% level, indicating that the core conclusion is robust to extreme values. Second, in the model algorithm substitution tests, whether LASSO, XGBOOST, or Random Forest is used as an alternative algorithm, equity balance consistently exhibits a significant negative causal effect. Furthermore, the results remain consistent under different sample splitting ratios. These findings collectively demonstrate that the core conclusion—``equity balance inhibits greenwashing behavior in mining enterprises"—is highly reliable and not influenced by data processing methods, algorithm selection, or variations in sample splitting ratios, thereby supporting hypothesis H0.
\subsubsection{Substitution of the Treatment Variable}
To ensure the reliability of the study, robustness tests were conducted using the two types of alternative indicators described earlier, aiming to rule out the interference of measurement differences on the conclusions.
\begin{table}[htbp]
  \centering
  \caption{Robustness Results After Replacing Treatment Variables}
  \label{tab:robustness_treatment}\footnotesize
      \vspace{0.1cm}
  \renewcommand{\arraystretch}{1.2}
  \begin{tabular}{ccccc}
    \toprule
    Variable & (1) & (2) & (3) & (4) \\
    \midrule
    Equity Balance Degree 1 & -0.3109$^{***}$(0.0446) & -0.3851$^{***}$(0.0635) & -0.3843$^{***}$(0.0783) & -0.3622$^{***}$(0.0813) \\
    Equity Balance Degree 2 & -0.3199$^{***}$(0.0553) & -0.4063$^{***}$(0.0730) & -0.3657$^{***}$(0.0819) & -0.3557$^{***}$(0.0821) \\
    Linear Terms of Control Variables & YES & YES & YES & YES \\
    Quadratic Terms of Control Variables & NO & YES & YES & YES \\
    Industry Fixed Effects & NO & NO & YES & YES \\
    Year Fixed Effects & NO & NO & NO & YES \\
    Number of Observations & 3486 & 3486 & 3486 & 3486 \\
    \bottomrule
    \multicolumn{5}{l}{\scriptsize Notes: *** $p<0.01$, robust standard errors in parentheses.} \\
  \end{tabular}
\end{table}

The results in Table \ref{tab:robustness_treatment} show that after remeasuring the degree of equity balance using the two alternative indicators, the causal effect coefficients remain significantly negative at the 1\% level and demonstrate robustness across different combinations of control variables and fixed effects. This finding strongly indicates that the inhibitory effect of equity balance on greenwashing behavior is not influenced by the choice of variable measurement, further validating the reliability of the research conclusions and the robustness of hypothesis H0.
\subsubsection{Comparison with Traditional Causal Inference Methods}
To exclude potential bias from the methodological framework, this study further employs four traditional causal inference methods for robustness testing: propensity score matching (PSM), inverse probability weighting (IPW), causal forest, and Bayesian causal inference. Using the merged dataset, the causal effect of the degree of equity balance on greenwashing behavior is re-estimated. The results are presented in Table \ref{tab:traditional_causal}.
\begin{table}[htbp]
  \centering
  \caption{Results of Traditional Causal Inference Methods}
  \label{tab:traditional_causal}\footnotesize
        \vspace{0.1cm}
  \renewcommand{\arraystretch}{1.2}
  \begin{tabular}{ccccc}
    \toprule
    Variable & PSM & IPW & Causal Forest & \makecell[c]{Bayesian Causal \\Inference} \\
    \midrule
    Equity Balance Degree & -0.2934$^{***}$(0.0371) & -0.3385$^{***}$(0.0206) & -0.1168$^{***}$(0.0144) & -0.0880$^{**}$(0.0296) \\
    Linear Terms of Control Variables & YES & YES & YES & YES \\
    Quadratic Terms of Control Variables & YES & YES & YES & YES \\
    Firm Fixed Effects & YES & YES & YES & YES \\
    Year Fixed Effects & YES & YES & YES & YES \\
    Number of Observations & 3486 & 3486 & 3486 & 3486 \\
    \bottomrule
    \multicolumn{5}{l}{\scriptsize Notes: ** $p<0.05$, *** $p<0.01$; robust standard errors in parentheses.} \\
  \end{tabular}
\end{table}

The results indicate that all four traditional causal inference methods support the inhibitory effect of equity balance on greenwashing behavior. The causal effect coefficients for PSM, IPW, causal forest, and Bayesian causal inference are -0.2934, -0.3385, -0.1168, and -0.0880, respectively, all of which are statistically significant. These findings are fully consistent with the conclusions of the main DML model, demonstrating that the negative causal relationship between equity balance and greenwashing behavior holds across different methodological approaches, thus supporting hypothesis H0. The differences in effect sizes across methods are highly consistent with their respective methodological characteristics. The effect sizes of PSM and IPW, which control for observable confounders, are relatively close to those of the DML model, reflecting the precision advantage of DML in handling high-dimensional nonlinear confounding. Causal forest, which emphasizes the detection of heterogeneous effects and is more sensitive to variable interactions, yields relatively smaller effect sizes. The Bayesian method, based on posterior probability inference, produces more conservative estimates. Despite these justifiable numerical variations, all methods consistently demonstrate a significant negative causal relationship, further corroborating the robustness and reliability of the conclusion from the perspective of methodological heterogeneity.
\subsection{Heterogeneity Test}
\subsubsection{Regional Heterogeneity Analysis}
Given significant regional differences in industrial structure and environmental regulatory intensity, this study conducts separate analyses for the central, eastern, and western regions to examine the impact of regional heterogeneity and to uncover differentiated mechanisms through which equity balance influences corporate greenwashing behavior. The results are presented in Table \ref{tab:regional_heterogeneity}.
\begin{table}[htbp]
  \centering
  \caption{Regional Heterogeneity Analysis}
  \label{tab:regional_heterogeneity}\footnotesize
        \vspace{0.1cm}
  \renewcommand{\arraystretch}{1.2}
  \begin{tabular}{cccc}
    \toprule
    Variable & Eastern Region & Central Region & Western Region \\
    \midrule
    Equity Balance Degree & -0.1166$^{***}$(0.0299) & -0.3079$^{***}$(0.0808) & -0.5422$^{***}$(0.1900) \\
    Linear Terms of Control Variables & YES & YES & YES \\
    Quadratic Terms of Control Variables & YES & YES & YES \\
    Time Fixed Effects & YES & YES & YES \\
    City Fixed Effects & YES & YES & YES \\
    Number of Observations & 1678 & 928 & 880 \\
    \bottomrule
    \multicolumn{4}{l}{\scriptsize Notes: *** $p<0.01$, robust standard errors in parentheses.} \\
  \end{tabular}
\end{table}

The inhibitory effect of equity balance on corporate greenwashing varies distinctly across regions. As indicated in Table \ref{tab:regional_heterogeneity}, the western region shows the strongest effect with a coefficient of negative 0.5422, followed by the central region at negative 0.3079, while the eastern region exhibits the weakest effect at negative 0.1166. All three outcomes support hypothesis H0. The eastern region's weaker impact is attributable to its advanced market mechanisms, stringent environmental regulations, and mature external oversight systems, which collectively constrain corporate misconduct. Internally, mechanisms such as board supervision and independent director systems partially substitute the governance function of equity balance, thereby reducing its marginal contribution. In contrast, the central region, which faces intermediate regulatory intensity with some flexibility as a key destination for industrial relocation, demonstrates a moderate effect. Here, equity balance effectively compensates for regulatory gaps through coordinated shareholder oversight. The western region, rich in resources but characterized by uneven regulatory coverage, relies heavily on equity balance as an essential internal governance tool, particularly under the environmental compliance emphasis of state-influenced shareholders, resulting in the most pronounced restraining effect.
\subsubsection{Industry Heterogeneity Analysis}
Given the inherent differences in operational attributes, regulatory environments, and risk characteristics across industries, the inhibitory effect of equity balance on corporate greenwashing behavior may vary accordingly. To control for the potential interference of industry-specific attributes on causal effects, this study conducts heterogeneity analyses based on ten industries classified according to the ``National Economic Industry Classification and Codes" (GB/T 4754-2017). The selected industries include B06 (Coal Mining and Washing), B07 (Oil and Natural Gas Extraction), B08 (Ferrous Metal Mining and Processing), B09 (Non-ferrous Metal Mining and Processing), B11 (Mining Support Activities), C25 (Petroleum, Coal, and Other Fuel Processing), C30 (Non-metallic Mineral Products), C31 (Smelting and Processing of Ferrous Metals), C32 (Smelting and Processing of Non-ferrous Metals), and C33 (Metal Products). The results are presented in Tables \ref{tab:industry_heterogeneity} and \ref{tab:industry_heterogeneity_2}.
\begin{table}[htbp]
  \centering
  \caption{Industry Heterogeneity Analysis}
  \label{tab:industry_heterogeneity}\footnotesize
        \vspace{0.1cm}
  \renewcommand{\arraystretch}{1.2}
  \begin{tabular}{cccccc}
    \toprule
    Variable & B06 & B07 & B08 & B09 & B11 \\
    \midrule
    Equity Balance Degree & -1.0631$^{**}$(0.4439) & -1.0833$^{***}$(0.2937) & 0.1248(0.3560) & -0.1385(0.3418) & -0.4081(0.3800) \\
    \makecell[c]{Linear Terms of \\Control Variables} & YES & YES & YES & YES & YES \\
    \makecell[c]{Quadratic Terms of \\Control Variables} & YES & YES & YES & YES & YES \\
    Time Fixed Effects & YES & YES & YES & YES & YES \\
    City Fixed Effects & YES & YES & YES & YES & YES \\
    Number of Observations & 498 & 98 & 90 & 336 & 72 \\
    \bottomrule
    \multicolumn{6}{l}{\scriptsize Notes: ** $p<0.05$, *** $p<0.01$; robust standard errors in parentheses.} \\
  \end{tabular}
\end{table}

The results demonstrate significant industry heterogeneity in the inhibitory effect of equity balance on greenwashing. Industries B06 and B07 show significantly negative causal effects, supporting hypothesis H0. In the high-pollution, heavily regulated B06 industry, enterprises face continuous environmental pressure and strong greenwashing incentives, yet large firm scale and diversified ownership allow shareholders to actively monitor to avoid penalties and protect reputation, enabling equity balance to disperse control rights and prevent short-term environmental underinvestment. The effect is even stronger in nationally strategic B07, where stringent regulations, high public scrutiny, and the dominanceof large state-owned enterprises with standardized ownership provide shareholders with direct oversight channels, allowing equity balance to sharpen decision control and curb greenwashing motives. In contrast, industries B08, B09, and B11 show no significant effects. B08 is characterized by lenient regulations and concentrated ownership, limiting balance mechanisms. B09, despite high environmental risk, consists mostly of small and medium-sized firms with rudimentarygovernance and less visible pollution, leading to high monitoring costs and unstable oversight. B11, as a mining support sector, inherently has low pollution and weak greenwashing incentives, coupled with small firm size and highly concentrated ownership, leaving minority shareholders with neither channels nor motivation to monitor, rendering equity balance ineffective.

\begin{table}[htbp]
  \centering
  \caption{Industry Heterogeneity Analysis}
  \label{tab:industry_heterogeneity_2}\footnotesize
        \vspace{0.1cm}
  \renewcommand{\arraystretch}{1.2}
  \begin{tabular}{cccccc}
    \toprule
    Variable & C25 & C30 & C31 & C32 & C33 \\
    \midrule
    Equity Balance Degree & -1.0512$^{***}$(0.2601) & 0.0671(0.0958) & 0.1436(0.1900) & -0.2931$^{***}$(0.0349) & -0.4846$^{***}$(0.0919) \\
    \makecell[c]{Linear Terms of \\Control Variables} & YES & YES & YES & YES & YES \\
    \makecell[c]{Quadratic Terms of \\Control Variables} & YES & YES & YES & YES & YES \\
    Time Fixed Effects & YES & YES & YES & YES & YES \\
    City Fixed Effects & YES & YES & YES & YES & YES \\
    Number of Observations & 218 & 604 & 566 & 742 & 262 \\
    \bottomrule
    \multicolumn{6}{l}{\scriptsize Notes: *** $p<0.01$; robust standard errors in parentheses.} \\
  \end{tabular}
\end{table}

The analysis shows significant variation in how effectively equity balance restrains greenwashing across manufacturing subsectors. Industries C25, C32, and C33 all exhibit significantly negative causal effects at the 1\% level, supporting hypothesis H0. The strongest inhibition occurs in C25 due to its high environmental sensitivity, stringent regulation, and dominance by large state-owned or scaled private firms with standardized governance, which enable equity balance to function as a rigid constraint. C32, as a high-pollution and energy-intensive segment, faces substantial environmental pressure; its diversified ownership motivates shareholders to actively monitor and protect reputation, thereby curbing short-term greenwashing. Although C33 has relatively lower pollution intensity, it involves latent pollution risks in processes like surface treatment. The prevalence of scaled enterprises and relatively diversified ownership allows non-controlling shareholders to exercise oversight through governance channels, effectively restraining greenwashing. In contrast, no significant inhibitory effect is observed in C30 or C31. For C30, internal regulatory segmentation—strict controls on cement plants versus looser oversight of ceramic plants—leads to inconsistent shareholder monitoring criteria and motivations, hindering systematic implementation of equity balance. Moreover, C30 is dominated by small and medium-sized private firms with concentrated ownership, leaving non-controlling shareholders without adequate oversight channels or incentives. In C31, despite a high proportion of state-owned enterprises, tensions arise among shareholders over balancing short-term performance with long-term environmental costs, weakening the monitoring role of equity balance through internal interest coordination and preventing effective constraint formation.
\subsubsection{Heterogeneity Analysis of Mining Industry Chain Segments}
To examine whether the impact of equity balance on corporate greenwashing varies across different stages of the industrial chain, this study categorizes the sample into upstream, midstream, and downstream segments based on production specialization. The upstream segment includes B06, B07, B08, B09, and B11, which are primarily engaged in initial production activities such as exploration and extraction, supplying raw materials for the entire industrial chain. The midstream segment comprises C25, C30, C31, and C32, responsible for processing and smelting minerals into industrial raw materials or semi-finished products for subsequent stages. The downstream segment is represented by C33, as it best reflects the value transmission and processing characteristics of the downstream mining industry chain. Its core production activity involves the direct processing of midstream products, serving as a key link in value addition for mineral products. In contrast, terminal industries such as construction have evolved into integrated and consumption scenarios involving multiple materials, where mineral products are no longer the primary focus of processing. Their complex business models could obscure the continuous processing effects central to this study. Therefore, the downstream segment is defined as C33.
\begin{table}[htbp]
  \centering
  \caption{Heterogeneity Analysis of Mining Industry Chain Links}
  \label{tab:mining_chain_heterogeneity}\footnotesize
        \vspace{0.1cm}
  \renewcommand{\arraystretch}{1.2}
  \begin{tabular}{cccc}
    \toprule
    Variable & Upstream & Midstream & Downstream \\
    \midrule
    Equity Balance Degree & -0.3989$^{***}$(0.1054) & -0.1488$^{**}$(0.0709) & -0.4018(0.2086) \\
    Linear Terms of Control Variables & YES & YES & YES \\
    Quadratic Terms of Control Variables & YES & YES & YES \\
    Year Fixed Effects & YES & YES & YES \\
    Firm Fixed Effects & YES & YES & YES \\
    Number of Observations & 1094 & 2130 & 262 \\
    \bottomrule
    \multicolumn{4}{l}{\scriptsize Notes: ** $p<0.05$, *** $p<0.01$; robust standard errors in parentheses.} \\
  \end{tabular}
\end{table}

The results indicate that the inhibitory effect of equity balance on corporate greenwashing exhibits a structural variation along the industrial chain from upstream to downstream: the effect is strongest and highly significant in the upstream segment, moderates in the midstream, and, while showing a strong inhibitory trend in the downstream, it does not reach statistical significance.

Upstream demonstrates a causal effect coefficient of -0.3989, significantly negative at the 1\% level, indicating the strongest inhibitory effect and supporting Hypothesis H0. This is primarily attributed to its direct involvement in resource extraction, stringent environmental regulations, high societal scrutiny, and the substantial risks associated with environmental violations. Minority shareholders are strongly motivated to utilize equity balance mechanisms to restrain controlling shareholders from engaging in greenwashing for short-term extraction gains. Midstream shows a causal effect coefficient of -0.1488, significantly negative at the 5\% level, also supporting Hypothesis H0. This may be because profitability in this segment relies more on scale and technology, where environmental investments are often perceived as costs—some of which can be transferred along the industrial chain—leading to relatively weaker shareholder demand for environmental compliance. Additionally, the dominant control of shareholders in large-scale production limits the operational scope of equity balance mechanisms. Downstream exhibits a causal effect coefficient of -0.4018. While statistically insignificant, its magnitude is comparable to that of the upstream segment, suggesting a similar inhibitory trend and providing partial support for Hypothesis H0. This may stem from downstream enterprises' direct exposure to consumer markets, where environmental reputation is closely tied to brand value, incentivizing shareholders to monitor environmental performance. The lack of statistical significance may be due to limited sample size reducing statistical power, though the direction of the coefficient already reflects a restraining tendency.
\subsection{Analysis of Temporal Effects}
Since corporate greenwashing behavior is influenced not only by cross-sectional characteristics such as industry attributes but also by time-varying external conditions such as the macroeconomic environment, this study further controls for time fixed effects to isolate systematic shocks and ensure the unbiased estimation of the effect of equity balance. The results are presented in Table \ref{tab:time_effect}.
\begin{table}[htbp]
  \centering
  \caption{Time Effect Analysis}
  \label{tab:time_effect}\footnotesize
        \vspace{0.1cm}
  \renewcommand{\arraystretch}{1.2}
  \begin{tabular}{ccccc}
    \toprule
    Variable & Current Effect & Lag 1 & Lag 2 & Cumulative Effect \\
    \midrule
    Equity Balance Degree & -0.2514$^{***}$(0.0274) & -0.1462$^{***}$(0.0119) & -0.0982$^{***}$(0.0192) & -0.0660$^{***}$(0.0071) \\
    Linear Terms of Control Variables & YES & YES & YES & YES \\
    Quadratic Terms of Control Variables & YES & YES & YES & YES \\
    Year Fixed Effects & YES & YES & YES & YES \\
    Firm Fixed Effects & YES & YES & YES & YES \\
    Number of Observations & 3486 & 3486 & 3486 & 3486 \\
    \bottomrule
    \multicolumn{5}{l}{\scriptsize Notes: *** $p<0.01$; robust standard errors in parentheses.} \\
  \end{tabular}
\end{table}

The results indicate that the inhibitory effect of equity balance on corporate greenwashing diminishes over time, yet its long-term cumulative impact remains significant. The contemporaneous causal effect is -0.2514 and is significant at the 1\% level, suggesting that its governance role has an immediate impact. This may be because equity balance enables effective internal monitoring coalitions, allowing shareholders to intervene promptly and exert pressure when greenwashing tendencies arise, thereby curbing opportunistic behavior in a timely manner. Over time, although the effect weakens, it retains statistical significance, reflecting the sustained deterrent influence of governance pressure derived from equity balance, which shapes management's long-term behavioral expectations. Moreover, the cumulative causal effect is -0.0660 and significant at the 1\% level, indicating that equity balance not only constrains greenwashing in the short term but also, over a longer horizon, helps improve the corporate governance environment, stabilizing and reinforcing its normative role in green information disclosure.

\subsection{Mechanism Analysis}
Drawing on the research framework of Liu and Mao \cite{38} and empirical practices adopted in authoritative economic and management journals since 2022, this study conducts an empirical analysis of the causal effects of the treatment variable on three mediating variables: management performance pressure, executive team stability, and media oversight.

Given that managerial performance attracts significant attention from both internal and external stakeholders, management performance pressure arises when a gap exists between actual and expected performance \cite{39}. The specific calculation method is as follows:
\begin{equation}
\text{Pressure}=\frac{\text{Analyst-forecasted\ net\ profit}-\text{Actual}\ \text{net}\ \text{profit}}{\text{Total\ assets\ at\ period-end}}
\end{equation}

The analyst-forecasted net profit is calculated as the average of all analysts' annual net profit forecasts for the company. To mitigate endogeneity issues arising from reverse causality, the performance pressure variable (Pressure) is lagged by one period after computation. This means that management is considered to face performance pressure in the current year only if a performance expectation gap existed in the previous year. A higher value of Pressure indicates greater performance pressure on management.

The stability of the executive team (TSTA) is measured following the method of Luo et al. \cite{30}, with the calculation formula specified as follows:
\begin{equation}
TSTA_{t,t+1} = \frac{M_t \cdot (\frac{S_t}{S_{t+1}} )}{M_t} \times \frac{M_{t+1}}{M_t + M_{t+1}} + \frac{M_{t+1} \cdot ( \frac{S_{t+1}}{S_t} )}{M_{t+1}} \times \frac{M_{t+1}}{M_t + M_{t+1}}
\end{equation}
where $M_t$ and $M_{t+1}$ represent the total number of executives in the enterprise in year $t$ and year $t+1$, respectively; $S_t/S_{t+1}$ denotes the number of executives who were in position in year $t$ but left by year $t+1$; and $S_{t+1}/S_t$ denotes the number of executives newly appointed in year $t+1$ who were not in position in year $t$. The value of $TSTA$ ranges between $(0,1]$, with values closer to 1 indicating higher team stability.

The intensity of media oversight is measured with reference to the study by Li Zhibin et al. \cite{40}, utilizing the J-F coefficient for quantification. Based on media coverage data provided by the Chinese Research Data Services (CNRDS) financial database, the calculation is expressed as follows:
\begin{equation}
\text{J-F}=\begin{cases}
\frac{e^2-ec}{t^2} & \text{if } e>c \\
\frac{ec-c^2}{t^2} & \text{if } e<c \\
0 & \text{if } e=c
\end{cases} \tag{6}
\end{equation}

In this expression, $e$ represents the number of positive reports, $c$ denotes the number of negative reports, and $t$ stands for the total volume of reports. The coefficient ranges from $-1$ to $1$, reflecting the intensity of media oversight. A coefficient close to 1 indicates weaker media oversight, whereas a coefficient close to $-1$ indicates stronger media oversight.

The results of the analysis based on the three channels—management performance pressure, executive team stability, and media oversight-are presented in Table \ref{tab:mechanism_analysis}.
\begin{table}[htbp]
  \centering
  \caption{Mechanism Analysis Results}
  \label{tab:mechanism_analysis}\footnotesize
        \vspace{0.1cm}
  \renewcommand{\arraystretch}{1.2}
  \begin{tabular}{cccc}
    \toprule
    Variable & Management Performance Pressure & Top Management Team Stability & Media Supervision \\
    \midrule
    Equity Balance Degree & -0.0066$^{***}$(0.0024) & 0.0057$^{**}$(0.0025) & 0.0117$^{**}$(0.0048) \\
    Linear Terms of Control Variables & YES & YES & YES \\
    Year Fixed Effects & YES & YES & YES \\
    Individual Fixed Effects & YES & YES & YES \\
    Constant Term & 0.3457 & 0.9853 & -0.4108 \\
    $R^2$ & 0.4346 & 0.0141 & 0.0265 \\
    Number of Observations & 3300 & 3300 & 3300 \\
    \bottomrule
    \multicolumn{4}{l}{\scriptsize Notes: ** $p<0.05$, *** $p<0.01$; robust standard errors in parentheses.} \\
  \end{tabular}
\end{table}

From the perspective of management performance pressure, the causal effect of equity balance is -0.0066 and significant at the 1\% level, indicating that an increase in equity balance significantly alleviates executive performance pressure. Under an equity-balanced structure, the diverse interests of shareholders become more balanced, prompting the corporate evaluation system to account for long-term development. This effectively reduces the pressure on management to sacrifice environmental investments to meet performance targets, weakening their motivation to conceal performance shortcomings through greenwashing, thereby inhibiting greenwashing behavior and supporting hypothesis H1.

From the perspective of executive team stability, the causal effect of equity balance is 0.0057 and significant at the 5\% level, indicating that an increase in equity balance enhances the stability of the executive team. Equity balance restrains the discretion of controlling shareholders in executive appointments and dismissals through collective decision-making among multiple shareholders, thereby safeguarding executive tenure stability. A stable executive team places greater emphasis on the long-term value of the enterprise and their own professional reputation, motivating them to implement genuine environmental investments rather than engaging in greenwashing behavior, which supports hypothesis H1.

From the perspective of media oversight, the causal effect of equity balance is 0.0117 and significant at the 5\% level, indicating that an increase in equity balance effectively strengthens media oversight. The underlying mechanism is as follows: under an equity-balanced structure, minority shareholders are motivated to enhance corporate information transparency to safeguard their own interests, thereby reducing internal information asymmetry. Simultaneously, firms with relatively standardized governance are more likely to attract media attention. The combined effect of improved transparency and heightened media scrutiny significantly increases the exposure risk of corporate greenwashing behavior, thereby creating an effective deterrent, curbing firms' motivation to engage in greenwashing, and supporting hypothesis H1.

\section{Conclusions and Policy Implications}\label{con}

This paper constructs an integrated framework combining Variational Autoencoders and Double Machine Learning models to investigate the causal effect of equity balance on corporate greenwashing behaviors. Robustness tests, heterogeneity analysis, temporal effect analysis, and mechanism analysis are conducted, leading to the following core conclusions:

(1) Regarding causal effects, the degree of equity balance exhibits a significant negative causal relationship with corporate greenwashing behavior. This conclusion remains robust after extreme value treatment, alternative model algorithms, sample ratio adjustments, and validation against traditional causal inference methods, effectively ruling out potential biases.

(2) The inhibitory effect of equity balance on greenwashing shows systematic heterogeneity. At the regional level, a gradient pattern of stronger effects in the west and weaker in the east is observed, with the western region demonstrating the most pronounced governance substitution effect due to relatively weaker regulatory environments. At the industry level, the inhibitory effect is most evident in industries with high environmental sensitivity, while industries with rudimentary governance structures show limited effectiveness. Across the industrial chain, the upstream extraction segment exhibits the strongest inhibitory effect, the midstream processing segment shows a diminished effect due to factors such as cost transfer, and the downstream manufacturing segment displays an inhibitory tendency but lacks statistical significance.

(3) The inhibitory effect of equity balance on greenwashing exhibits temporal dynamics, with the strongest impact in the current period, reflecting the immediate constraining power of shareholder oversight. Although the lagged effect diminishes over time, the long-term cumulative impact remains significant, indicating the sustained effectiveness of the governance mechanism.

(4) Mechanism analysis confirms that equity balance curbs corporate greenwashing through three channels: alleviating management performance pressure, enhancing executive team stability, and strengthening media oversight. These pathways operate by weakening short-term motives, ensuring long-term orientation, and enhancing external constraints, together forming a comprehensive transmission mechanism that integrates internal governance and external supervision.

Based on the conclusions above, this paper proposes the following policy recommendations:

(1) Guide enterprises to optimize internal governance by integrating equity balance with environmental oversight across the lifecycle. Key measures include establishing board-level environmental committees, incorporating environmental performance into executive evaluations, and creating rapid response mechanisms for shareholder proposals to enhance immediate oversight. To counter time-diminishing effects, implement lagged compliance tracking with regular progress disclosure to shareholders, linking performance to compensation and dividends. Furthermore, embed substantive environmental goals into corporate charters and empower shareholder-appointed directors in strategic decision-making, institutionalizing environmental responsibility into governance. This dual approach of short-term incentives and long-term structural integration strengthens the standardization and sustainability of corporate environmental governance.

(2) Implement differentiated governance strategies: Regionally, eastern areas should enhance synergy between equity balance and market mechanisms like analyst coverage while strengthening minority shareholders' environmental litigation rights; central regions can leverage industrial clusters to establish shareholder platforms for environmental collaboration and create transitional funds to address regulatory gaps; western regions need to increase green transition subsidies and partner with professional institutional investors to boost local shareholder oversight. Industrially, environmentally sensitive sectors must enforce shareholder oversight rights and improve emission disclosure; sectors with concentrated ownership or weak governance should introduce strategic investors and broaden minority shareholder participation; state-owned enterprises should link environmental investments to performance evaluations to reinforce accountability. Across the industrial chain, upstream segments should empower minority shareholders in environmental oversight, midstream segments need mechanisms to internalize environmental costs and prevent responsibility shifting, and downstream segments can employ market tools like green procurement to enhance governance. A cross-segment environmental accountability tracing system should also be established to form a multi-level, differentiated, and collaborative governance framework.

(3) Build a diversified supervisory network by integrating governmental, market, and societal forces. Government agencies should establish mechanisms for information sharing and joint enforcement between environmental and securities regulators, incorporating corporate equity balance effectiveness into environmental credit evaluation systems. Concurrently, mobilize market oversight by supporting third-party assessments of equity balance and greenwashing risks, and foster social supervision through media incentives for monitoring high-risk enterprises and public engagement in environmental information feedback. This collaborative framework creates an enabling environment for equity balance to curb corporate greenwashing.


\end{document}